%% file: paper_stabledpt.tex
\documentclass[10pt,twocolumn,letterpaper]{article}

\usepackage[pagenumbers]{cvpr} %

\input{preamble}
\usepackage{xcolor}

\definecolor{cvprblue}{rgb}{0.21,0.49,0.74}
\usepackage[pagebackref,breaklinks,colorlinks,allcolors=cvprblue]{hyperref}

\def\paperID{REDACTED} %
\def\confName{CVPR}
\def\confYear{2026}

\title{StableDPT: Temporal Stable Monocular Video Depth Estimation}

\author{Ivan Sobko$^{1, 2}$
\hspace{7mm}
Hayko Riemenschneider$^{2}$
\hspace{7mm}
Markus Gross$^{1, 2}$
\hspace{7mm}
Christopher Schroers$^{2}$
\\
$^{1}$ETH Z\"urich
\hspace{7mm}
$^{2}$DisneyResearch\textbar Studios \\
{\tt\small {isobko@ethz.ch}} \hspace{7mm} {\tt\small {grossm@inf.ethz.ch}} \\ 
{\tt\small {\{hayko.riemenschneider,christopher.schroers\}@disneyresearch.com}}
}

\begin{document}

\input{sec/0_abstract}
\input{sec/1_intro}

\input{sec/2_relatedwork}

\input{sec/3_method}

\input{sec/4_experiments}

\input{sec/6_conclusion}

\clearpage

{
    \small
    \bibliographystyle{ieeenat_fullname}
    \bibliography{main}
}

\end{document}

%% file: sec/0_abstract.tex
\twocolumn[{
            \maketitle %
            \begin{center}
                \includegraphics[width=0.80\textwidth]{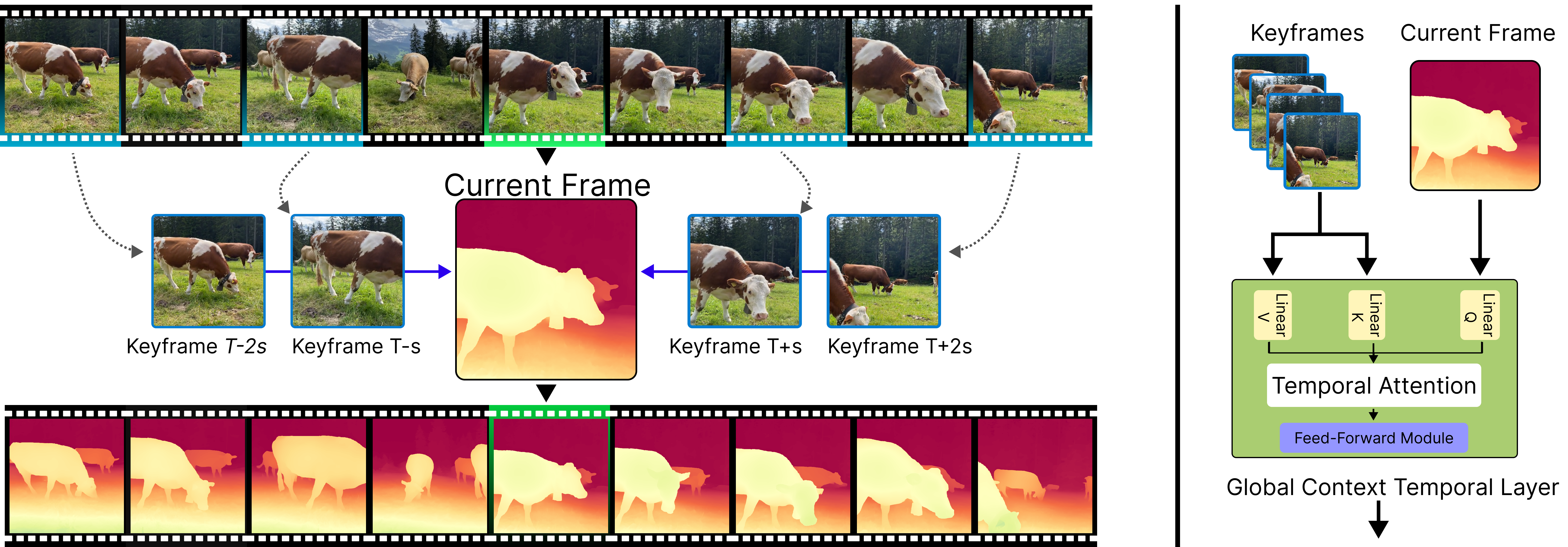}
                \captionof{figure}{Left: Our model, StableDPT, can leverage information from the entire video sequence to produce temporally stable and accurate depth maps from monocular video input with complex camera and scene dynamics.
                    Right: our keyframes can effectively capture global context of the video, and anchor current predictions via temporal module with cross-attention.
                }
                \label{fig:teaser}
            \end{center}
        }]

\begin{abstract}
Applying single image Monocular Depth Estimation (MDE) models to video sequences introduces significant temporal instability and flickering artifacts.
We propose a novel approach that adapts any state-of-the-art image-based (depth) estimation model for video processing by integrating a new temporal module - trainable on a single GPU in a few days.

Our architecture StableDPT builds upon an off-the-shelf Vision Transformer~(ViT) encoder and enhances the Dense Prediction Transformer~(DPT) head.
The core of our contribution lies in the temporal layers within the head, which use an efficient cross-attention mechanism to integrate information from keyframes sampled across the entire video sequence.
This allows the model to capture global context and inter-frame relationships leading to more accurate and temporally stable depth predictions.

Furthermore, we propose a novel inference strategy for processing videos of arbitrary length avoiding the scale misalignment and redundant computations associated with overlapping windows used in other methods.
Evaluations on multiple benchmark datasets demonstrate improved temporal consistency, competitive state-of-the-art performance and on top 2x faster processing in real-world scenarios.
\end{abstract}

%% file: sec/1_intro.tex
\section{Introduction}

Depth estimation is the fundamental computer vision problem to predict the distance from a camera to scene objects.

The deep learning revolution has impacted the depth estimation field, enabling prediction from a single image, known as Monocular Depth Estimation (MDE).
MDE can now be performed using standard cameras, which are widely available and cost-effective, making depth estimation more accessible.

However, despite advancements, the inherent ambiguity makes prediction challenging, since information from a single image
is often insufficient to correctly infer the underlying depth structure.
Many existing methods struggle with generalization across different scenes~\cite{eigen2014depthmappredictionsingle,roy2016CVPR} or
require significant computing power, such as (video) diffusion methods~\cite{ke2025marigold,hu2024DepthCrafter},
making them difficult to use in real-time applications or on devices with limited resources.

While recent methods~\cite{ranftl2022Midas, yang2024DepthAnything,yang2024DepthAnythingV2,bochkovskii2024DepthPro} have
achieved impressive results on single image depth estimation, they are not designed to utilize temporal information present in video sequences.
As a result, when applied to video data, these approaches often produce depth maps with flickering artifacts and blurry results.

In this paper, we propose a novel approach that leverages temporal information from video sequences to improve depth prediction
without using any supplementary sensors or data.
Our model, StableDPT, introduces a new way to utilize a temporal module that processes multiple frames in a video sequence,
allowing the model to capture inter-frame relations and predict more accurate and stable depth.

Our temporal module consists of a transformer-based architecture with cross-attention mechanisms that enable the
model to focus on relevant features from the relevant frames.
We evaluate our method on several benchmark datasets, demonstrating its effectiveness in producing temporally consistent depth maps
that match state-of-the-art methods - at more efficient inference and avoiding heuristic post-processing schemes.
In summary, our method enables precise and temporally consistent depth estimation:

\begin{itemize}
  \item Global efficient temporal cross-attention for adapting image-based models to stable video prediction.
  \item Efficient inference strategy for processing videos of arbitrary length, avoiding scale misalignment and redundant computations.
  \item Competitive state-of-the-art performance in accuracy and temporal consistency for real-world scenarios.
\end{itemize}

%% file: sec/2_relatedwork.tex
\section{Related work}

Monocular depth prediction~(MDE) research has evolved in multiple directions including single image depth, guided control points
depth, multi-view reconstruction, video depth and video diffusion-based depth. Each direction has its own merits and challenges.
However, our focus lies on efficient temporally stable video depth estimation methods without relying on additional sensors or data.

\subsection{Single image prediction}

To this day, monocular depth estimation remains a challenging task due to the inherent ambiguity in inferring depth from a
single image. Current single image estimation approaches can be broadly categorized into two main types: transformer-based and
diffusion-based methods.

The introduction of MiDaS~\cite{ranftl2022Midas} marked a significant breakthrough in monocular depth estimation, as it was the
first to use mixed datasets to enhance generalization across different scenes and conditions. After MiDaS, several other
methods emerged, such as DepthAnything (V1, V2)~\cite{yang2024DepthAnything, yang2024DepthAnythingV2} using
DINOv2~\cite{oquab2023dinov2} backbone, which proposed to train large-scale model on precise, synthetic data and then use this
model as a teacher for training a smaller model on unlabeled data, thus achieving even better generalization.
DepthPro~\cite{bochkovskii2024DepthPro}, also using a DINOv2 backbone, improves the depth sharpness by introducing a new
training strategy that scales image and then merges features from different scales.

Another method stems from the field of image diffusion, where the depth prediction is modeled as a generative process.
Marigold~\cite{ke2025marigold} and follow-up methods~\cite{talegaonkar2025repurposingmarigoldzeroshotmetric, he2024lotus,
fu2024geowizard} use rich visual knowledge from modern generative models~\cite{rombach2022stablediffusion} to predict zero-shot data,
or a better combination~\cite{zhang2024betterdepth}.

\subsection{Multi-view depth reconstruction}
Structure from Motion is a foundational methodology for recovering 3D structure of a static scene from a set of images
captured from unknown viewpoints. The process simultaneously estimates a sparse 3D point cloud of the scene and the
six-degree-of-freedom (6~DoF) pose for each camera. While capable of achieving high accuracy, it can be computationally
intensive and sensitive to challenges such as textureless surfaces and dynamic scenes, motivating the development of
learning-based alternatives. The Visual Geometry Grounded Transformer (VGGT)~\cite{wang2025vggt} represents a shift from
classical 3D reconstruction towards a direct, feed-forward inference approach. It introduces a large transformer architecture
capable of processing unstructured views of a scene to jointly infer a comprehensive set of 3D attributes,
including camera intrinsics, extrinsics, dense depth maps, 3D point maps, and point tracks.

\subsection{Consistent depth estimation}
Recent advancements in video depth estimation have focused on using inter-frame relations to improve the consistency across
frames.
Recent methods such as VideoDepthAnything~\cite{chen2025videodepthanything}, DepthCrafter~\cite{hu2024DepthCrafter},
RollingDepth~\cite{ke2024rollingdepth} and FlashDepth~\cite{chou2025flashdepth} leverage transformer architectures with self-attention
mechanisms to capture temporal dependencies across videos.

\paragraph{VideoDepthAnything}~(VDA) builds upon DepthAnythingV2 and the Dense Prediction Transformers~(DPT)
head~\cite{ranftl2021DPT}, introducing temporal layers from AnimateDiff~\cite{guo2023animatediff}. 
These layers incorporate self-attention~\cite{vaswani2023attentionneed} and feed-forward networks that operate exclusively in the temporal dimension, facilitating effective interaction of temporal features.
Additionally, VDA introduces a novel Temporal Gradient Matching~(TGM) loss function, which is based on the Optical-based Flow Warping~(OPW) loss~\cite{wang2023NVDS}. 
Temporal stability in VDA is further enhanced by a two-stage inference strategy: In the first stage, the video is processed in overlapping windows, with each window
including two frames from the previous window to maintain global information across windows. 
In the second stage, these windows are stitched together and the overlapping regions are interpolated to ensure smooth transitions. 
While this inference strategy contributes to video stability, the smoothing can cause data loss, particularly in overlapping regions.

\paragraph{RollingDepth} adapts a single image diffusion model for video depth estimation, such as Marigold~\cite{ke2025marigold}, by adjusting its
self-attention layers to handle short video snippets. Video snippets are created by sampling many overlapping snippets using a
\textit{dilated rolling kernel} to capture both short and long-range dependencies. Temporal coherence is achieved in two steps: First, the self-attention
layers are modified to operate across all frames in a snippet. Second, a global, optimization-based co-alignment step is
performed at inference time. It calculates the optimal scale and shift for every snippet to minimize the L1 difference on
overlapping frames, ensuring long-range consistency. For the post-processing, RollingDepth takes the co-aligned depth video,
adds a moderate amount of noise, and denoises it again using the same model to enhance fine details.

\paragraph{DepthCrafter} uses a pre-trained Stable Video Diffusion~(SVD)~\cite{blattmann2023SVD} model to generate temporally consistent and detailed
depth maps for long, open-world videos. Temporal consistency is primarily inherited from the inherent temporal layers and
motion priors of the SVD model and three stage training strategy, which includes training on hand-crafted dataset mixture with
synthetic and real-life data and a specific stage for temporal layer's fine-tuning on variable video lengths. The resulting
model processes video in segments, with a maximum length of 110 frames at a time. For longer videos, it uses a segment-wise
inference strategy where the video is broken into overlapping chunks.

\paragraph{FlashDepth} is another approach to video stabilization is based on utilizing recurrent neural networks, to capture and transfer hidden
state from frame to frame. FlashDepth focuses on achieving real-time, high-resolution (2K) depth
estimation for streaming video inputs. It is based on DepthAnythingV2 and uses structured state-space model~(SSM)~\cite{gu2022efficiently}, 
particularly Mamba~\cite{gu2024Mamba}, which operates on intermediate features from the model's
decoder, aligning the depth scale of each frame on-the-fly by updating a hidden state that carries information from past
frames. Using SSMs allows processing videos in a streaming fashion, without the need to see future frames and without retaining
much prior knowledge. FlashDepth has no special loss for temporal consistency and no post-processing, relying
solely on the Mamba module to ensure temporal coherence. The results, however, indicate that Mamba alone is not sufficient to fully address temporal coherence.

\subsection{Video transformers}

Generally, our approach is similar by Video Transformer architectures~\cite{selva2022videotransformersurvey}.
These architectures extend the transformer architecture to handle the spatio-temporal nature of video data.
They have been successfully applied to various video understanding tasks, such as action recognition~\cite{xiao2025trajectory} and video understanding~\cite{feichtenhofer2018slowfast}.
However, our method differs in that instead of classification it specifically targets dense output prediction tasks, \eg~monocular depth estimation.

Video Swin Transformer~\cite{liu2021video} extends the Swin Transformers~\cite{liu2021swin,liu2022swinv2} to video by incorporating sliding 3D windows for spatiotemporal modeling.
This is a very dense but local representation and does not capture global context as effectively as our cross-attention mechanism.

Similar to the taxonomy introduced in TimeSformer~\cite{bertasius2021timesformer}, we follow a divided space-time attention scheme.
However, our focus is on dense prediction for temporal stability and improved accuracy.

The ViViT~\cite{arnab2021vivit} architecture introduces a factorized attention mechanism that separates spatial and temporal attention.
This allows the model to focus on relevant features in each domain independently, improving its ability to efficiently capture complex spatiotemporal relationships.
They encode each frame separately using a pre-trained spatial Vision Transformer (ViT)~\cite{dosovitskiy2020vit} and then apply a temporal transformer to the sequence of frame embeddings.
In our work, we also extract spatial features yet directly cross-attend to them with interleaved temporal and spatial attention.

%% file: sec/3_method.tex
\section{Method: StableDPT}
\label{sec:method}

\begin{figure*}[t]
    \centering
    \includegraphics[width=0.9\linewidth]{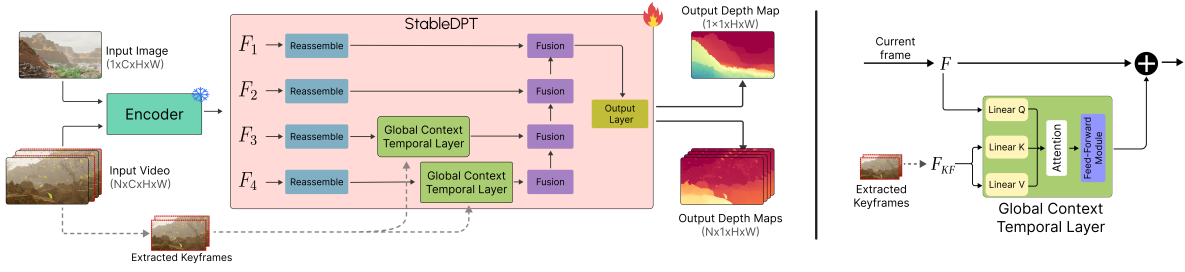}
    \caption{Overall architecture of our model. Left: The encoder extracts features from each frame independently, while the head
        processes these features with global context temporal layers that integrate information from keyframes sampled from the entire video sequence.
        The output is a depth map for each frame in the input video.
        Right: Detailed view of the temporal layer, which uses cross-attention mechanisms to align features with information from the global context.}
    \label{fig:method_diagram}
\end{figure*}

Our research addresses the temporal instability of Monocular Depth Estimation~(MDE) in videos by integrating temporal transformers into single image depth prediction.
Similar to the the seminal MiDaS~\cite{ranftl2022Midas} with its Dense Prediction Transformers~(DPT) head~\cite{ranftl2021DPT} for single image estimation,
and Video Depth Anything~(VDA) with its Spatial-Temporal Head~(STH) for video estimation, we propose a novel efficient architecture that works for both worlds.

Our method uses a global context build on keyframes, sampled from the entire video sequence.
This allows the model to leverage information from close and distant frames for more accurate and robust depth estimation.
Our investigations indicate that spatial information is adequately captured by the pre-trained image encoder.
Hence, temporal context is the missing piece for stable video prediction and the decoder head can solely focus on temporal modeling.

Our architecture follows the established encoder-decoder framework commonly used in depth estimation.
We exploit this division and create a temporal cross-attention mechanism in the decoder as illustrated in Figure~\ref{fig:method_diagram}.
Our decoder is a temporal-only model by design that focuses on integrating temporal information from the global context keyframes, see Figure~\ref{fig:attention_viz}.
Prior work like VDA processes video in overlapping windows, incorporating only two previous keyframes for temporal context and attends only to frames within the batch.
They employ a heavy post-processing scheme involving moving keyframes and overlapping frames to be able to mitigate scale misalignment and temporal coherence.

StableDPT avoids this entirely and can be trained end-to-end without any post-processing or alignment.
Our temporal global context enables StableDPT to better capture information from the entire video sequence directly producing stable aligned depth maps.

\subsection{Encoder Architecture}

As with MiDaS~\cite{ranftl2022Midas} and DPT~\cite{ranftl2021DPT}, we use an architecture similar to ViT~\cite{dosovitskiy2020vit} as the encoder, which consists of a series of transformer layers that process the input image as a sequence of patches.
However, StableDPT is agnostic to the choice of encoder and can work with any ViT-based architecture.

Each patch is treated as a token, and the transformer layers learn to capture relationships between these tokens to extract meaningful spatial features from the independent image, as illustrated in Figure~\ref{fig:attention_viz}.
The encoder takes a sequence of video frames as input images, denoted $ I \subset \mathbb{R}^{N \times C \times H \times W} $,
where $N$ represents the length of the video sequence, $C$ is the channel count, and $H$ and $W$ are the height and width of the frames, respectively.
The encoder processes each image independently, extracting intermediate features from each frame using the same set of transformer layers,
denoted $ F_i \subset \mathbb{R}^{N \times C_i \times \frac{H}{p} \times \frac{W}{p}} $, where $p$ is the patch size of the encoder token.

The extracted features are then passed to the decoder head for further processing.
Please note that the encoder's spatial processing is sufficient, and the decoder subsequently treats each token as spatially independent.

\subsection{Decoder Architecture}

The head of our model is similar to a modified DPT architecture with insertion of temporal layers from VDA.
However, unlike VDA, which processes video in overlapping windows with inclusion of two previous frames, our approach samples keyframes
$I_{kf} = \{I_{kf1}, I_{kf2}, \ldots I_{kfM}\}$ from the entire video sequence and modifies the temporal layer to use cross-attention mechanisms
to integrate information from these keyframes into the current frame's features.

\begin{figure}[t]
    \centering
    \includegraphics[width=0.85\linewidth]{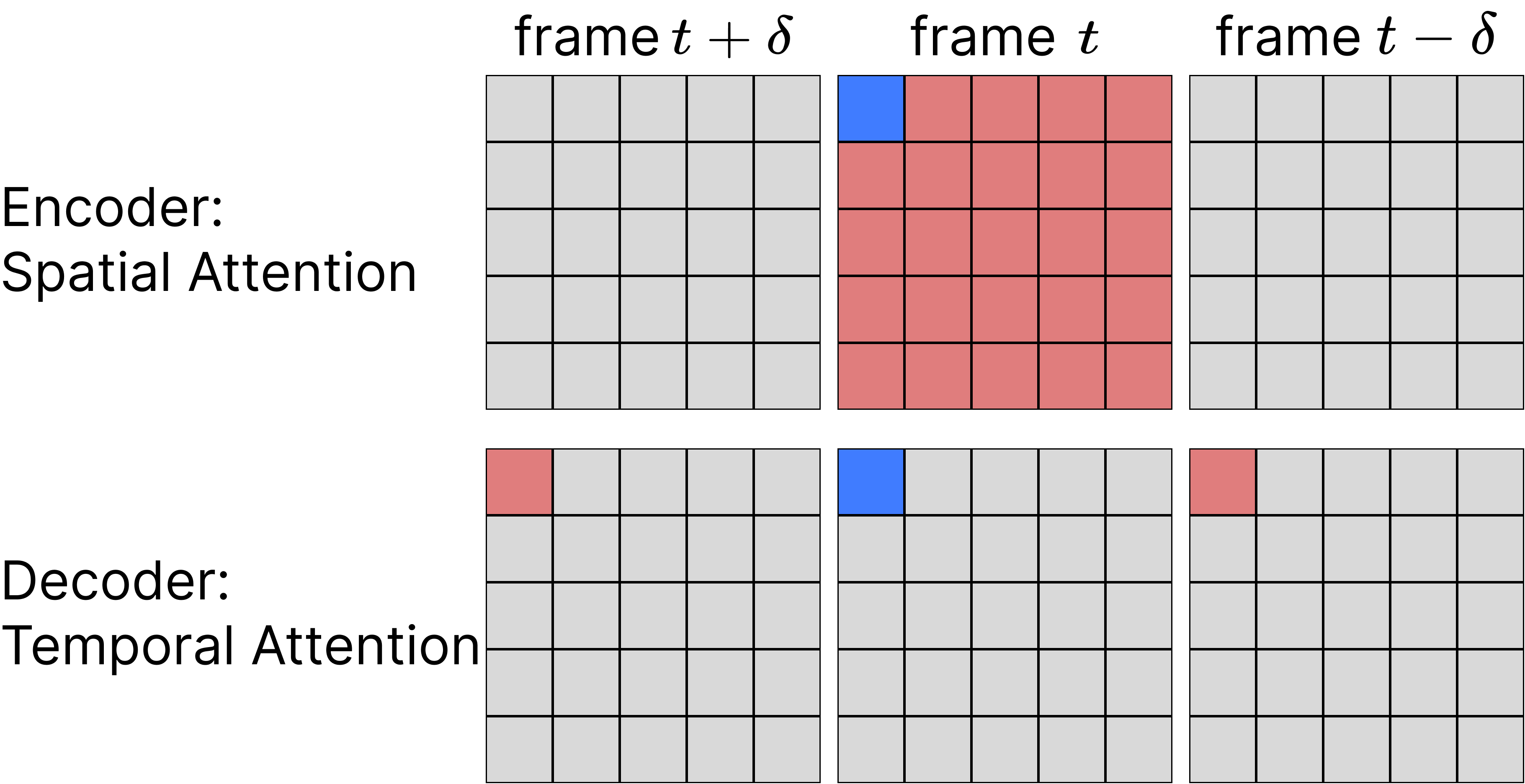}
    \caption{Visualization of attention maps from the encoder and decoder attention layers.
        Top row: Attention maps from the encoder show spatial attention within individual frames, capturing relationships between different regions of the same image.
        Bottom row: Attention maps from the decoder temporal layers illustrate attention across time, highlighting how the temporal layer informs current features about surrounding frames, past and future.}
    \label{fig:attention_viz}
\end{figure}

The DPT head takes the intermediate feature maps $ F_i $ from the encoder and passes them through a Reassemble layer, which reconstructs image-like representations from the token outputs of selected transformer encoder layers.
These representations are then reshaped and processed by temporal layers to align features across the video sequence.
The reshaping is designed in a way to restrict attention computation for each image patch across time only, as illustrated in Figure~\ref{fig:attention_viz}.
This restriction significantly reduces computational cost while still promoting awareness of temporal context for each processed frame.
Formally, given a set of keyframes $I_{kf}$, the cross-attention computes attention weights between the current frame's features and those of all keyframes:
\begin{equation}
    F_{att} = \mathrm{Att}(Q_{\mathrm{curr}}, K_{\mathrm{kf}}, V_{\mathrm{kf}}) = \mathrm{softmax}\left(\frac{Q K^\top}{\sqrt{d_k}}\right) V,
\end{equation}
where $Q_{curr}$ are the queries from the current frame, and keys $K_{kf}$ and values $V_{kf}$ are from the keyframes $I_{kf}$.

This allows the model to aggregate information from temporally distant frames and use context from the entire video,
rather than being limited to previous frames, resulting in more robust depth estimation.
The temporal layers consist of transformer layers with multi-head cross-attention mechanisms and a feed-forward network~(FFN).
We zero-initialize the last layer of the temporal layers to stabilize training and preserve original features at the start of training,
while gradually allowing the model to learn temporal dependencies.
To extract the most relevant temporal information, we insert temporal layers at multiple scales within the head.
Specifically, we add temporal layers after the Reassemble module on the two feature maps with the highest semantic content~(i.e.,
the deepest layers of the encoder) to maximize the amount of temporal alignment.

The output of the temporal layers is a set of aligned feature maps, denoted $ AF_i' \subset \mathbb{R}^{N \times C_i
    \times \frac{H}{p} \times \frac{W}{p}} $.
These layers are then fused using a Fusion module, which combines the multiscale features into a single feature map.
The fused feature map is then processed by a series of convolutional layers to refine the features before being upsampled to the original image resolution.
Both Reassemble and Fusion modules are similar to the corresponding modules from the original DPT paper~\cite{ranftl2021DPT}.

The output of the head is a depth map for each frame in the input video sequence, denoted $D \subset \mathbb{R}^{N \times 1 \times H \times W}$.

\subsection{Inference Strategy}
\label{sec:inference_strategy}

The elegance of single-frame depth estimation models lies in their ability to process images independently, allowing for efficient batch processing.
However, when extending these models to video sequences, challenges arise in maintaining temporal consistency, efficient processing, avoiding depth drift and handling varying video lengths.

\begin{figure}[t!]
    \centering
    \begin{subfigure}[c]{1.0\linewidth}
        \includegraphics[width=1.0\linewidth]{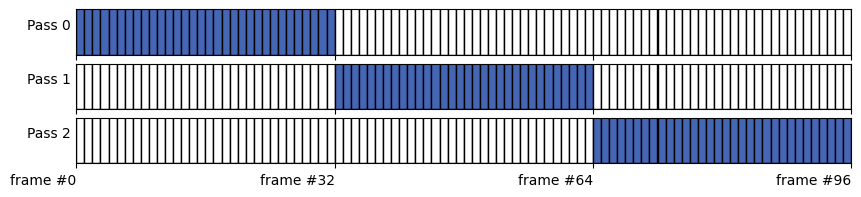}
        \subcaption{Batched inference}
        \label{fig:inference_batched}
    \end{subfigure}

    \begin{subfigure}[c]{1.0\linewidth}
        \includegraphics[width=1.0\linewidth]{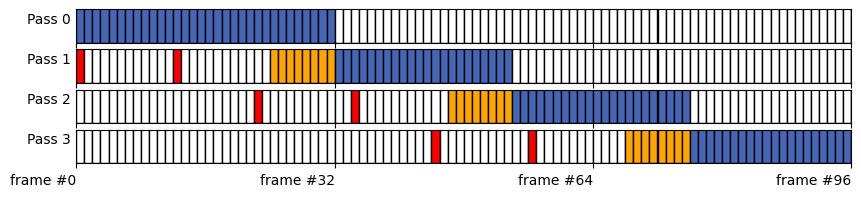}
        \subcaption{VDA inference (Orange boxes indicate overlapping regions)}
        \label{fig:inference_vda}
    \end{subfigure}

    \begin{subfigure}[c]{1.0\linewidth}
        \includegraphics[width=1.0\linewidth]{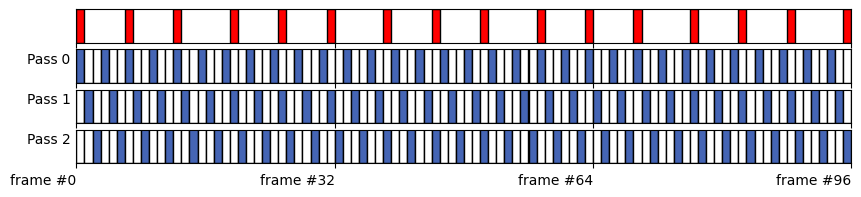}
        \subcaption{Strided inference (Red boxes indicate global keyframes)}
        \label{fig:inference_strided}
    \end{subfigure}

    \caption{Illustration of different inference strategies for processing videos of arbitrary length $N$.
        (a) Batched inference divides the video into non-overlapping batches of fixed length, leading to scale misalignment between batches.
        (b) VDA inference processes overlapping windows and interpolates the overlapping regions, which requires processing more frames.
        (c) Strided inference samples frames at regular intervals from the entire video, ensuring each frame is processed with context from the whole video.}
    \label{fig:inferences_strategies}
\end{figure}

To process videos of arbitrary length $N$, we evaluated several inference strategies.
The simplest approach divides the video into windows of fixed length (see Figure~\ref{fig:inference_batched}), but this leads to scale misalignment between windows.
VDA addresses this by processing overlapping windows, as shown in Figure~\ref{fig:inference_vda}.
The overlapping regions are then interpolated and aligned through affine transformation to ensure smooth transitions between windows.
They also utilize keyframing to anchor predictions to a consistent scale and reduce accumulated scale drift, innate to affine alignment.

However, this approach has several inherent limitations.
First, it only uses two previous keyframes for temporal context, which limits the model's ability to leverage information from the entire video sequence, past and future.
Second, overlapping windows require approx.~1.5x more computation than needed, due to overlapping frames being processed multiple times,
and introduces inconsistencies at the boundaries of the windows.

To overcome these limitations, we propose a strided sliding window approach.
As illustrated in Figure~\ref{fig:inference_strided}, our strategy creates snippets of length $ L_s $, each containing frames sampled from the entire video.
We sample frames at regular intervals, defined by a stride $s = \lfloor N / L_s \rfloor$, each snippet thus contains frames
$ \{ I_{i}, I_{i+s}, I_{i+2s}, \ldots, I_{i+(L_s-1)s} \} $ for $ i = 0, 1, \ldots, s-1 $.
Snippets are processed independently, with keyframes $I_{kf}$, acting like an anchor across all snippets.
Keyframes remain consistent across all snippets, ensuring that each snippet is grounded in the same temporal context - effectively anchoring it and avoiding scale drift.

Our benefit is twofold: First, by sampling frames from the entire video, we ensure that each snippet captures a broad temporal context,
allowing the model to leverage information from both past and future frames.
Second, by avoiding overlapping windows, we reduce computational redundancy and eliminate the need for interpolation and affine alignment, leading to more efficient processing.

The final depth maps are obtained by simple reorder of the outputs to match the original frame sequence.

\subsection{Training pipeline}
During training, we use a combination of image and video datasets.
Following \cite{yang2024DepthAnythingV2}, image data is processed with a teacher model to generate pseudo-ground truth depth maps, which are used for semi-supervised training.
As a teacher, we use the base DepthAnythingV2-large image model.

For video data, we use supervised training, leveraging temporal consistency between consecutive frames.
We alternate between image and video data during training to balance the two types of data: image datasets provide diverse scenes, while video datasets capture temporal dynamics.
In practice, both types of data are passed through the complete model, including the temporal layers.
Our experiments indicate that differentiating between images and videos during processing - \eg~bypassing temporal layers for images - does not yield any benefit.

We use a combination of supervised losses to train our model.
Similar to Ranftl et~al.~\cite{ranftl2022Midas}, we use the SSI Trim loss and the Gradient Matching loss, which encourage accurate depth predictions while being robust to outliers.

For temporal consistency, we use the Temporal Gradient Matching loss~\cite{chen2025videodepthanything},
which encourages smooth depth transitions between consecutive frames.
The loss is calculated on scale-shift aligned depths~$\hat{d}$ and ground truth~$\hat{d}'$ and is defined as:
\begin{equation}
    L_{TGM} (\hat{d}, \hat{d}') = \tfrac{1}{N-1} M \sum_{i=1}^{N-1} \left\| |\hat{d_{i+1}} - \hat{d_i}| - | \hat{d_{i+1}'} -
    \hat{d_i'}| \right\|_1
\end{equation}
where $M$ is a mask that excludes pixels with large intensity changes, which correspond to object boundaries, dynamic objects or occlusions.
The mask is defined as:
\begin{equation}
    M = \{1, \text{if } |I_{i+1} - I_i| < \tau; 0, \text{otherwise}\}
\end{equation}
where $\tau$ is a threshold value, which we set to $\tau=0.05$.

For image data, we only apply the SSI Trim loss and the Gradient Matching loss, while for video data, we use all three losses.
Our final loss function is a weighted combination of these losses with $\lambda_{ssi} = 2.0$, $\lambda_{gm} = 1.0$ and $\lambda_{tgm} = 2.5$:
\begin{equation}
    L = \lambda_{ssi} L_{SSI trim} + \lambda_{gm} L_{RG} + \lambda_{tgm} L_{TGM}.
\end{equation}

%% file: sec/4_experiments.tex
\section{Experiments}

We evaluate our proposed method on multiple benchmark datasets for video depth estimation, following the protocols used in previous works~\cite{ranftl2021DPT,ranftl2022Midas,chen2025videodepthanything,chou2025flashdepth}.

We utilize a pre-trained ViT encoder DepthAnythingV2~\cite{yang2024DepthAnythingV2}, which have been trained on large-scale image datasets and proved to be effective for depth estimation tasks.
During training, we keep the encoder frozen during video training to minimize computational requirements and retain the encoder's pre-trained representations.
As noted, any other ViT-based encoder can be used as well, since StableDPT is agnostic to the choice of encoder and task.

\subsection{Evaluation}

\paragraph{Datasets.}
We use a combination of real-world and synthetic datasets with different scenarios, different camera movements, and varying levels of complexity.
For training, since not all data from VDA is public, we use the following data: a synthetic urban environment dataset (20k), TartanAir~\cite{wang2020tartanair} (306k), Spring~\cite{mehl2023Spring} (5k) and unlabelled frames from movie data (29k), in total approximately 360k frames.
For evaluation, we use the commonly used benchmark datasets such as Sintel~\cite{butler2012sintel}, KITTI~\cite{geiger2012KITTI}, TUM RGB-D~\cite{palazzolo2019bonn} and the recent Infinigen~\cite{raistrick2023infinite}, which provides the most accurate and high-resolution ground truth depth (2560x1440) as used for evaluation.

\paragraph{Metrics.}
Since our method focuses on improving temporal consistency in video depth estimation, we evaluate our models using both image accuracy and temporal consistency metrics.
For accuracy, we use the standard metrics: Absolute Relative Error (AbsRel)~\cite{ranftl2022Midas} and $\delta_1$ accuracy~\cite{ranftl2022Midas}.
For temporal consistency, we use the TGM metric~\cite{chen2025videodepthanything}, Temporal Consistency (TC)~\cite{li2021enforcing}, Temporal Consistency Correlation (TCC)~\cite{khan2023tcod}, Temporal Motion Consistency (TMC)~\cite{khan2023tcod}, and Optical Flow Warping (OPW)~\cite{wang2023NVDS} metric.
These metrics provide a more comprehensive evaluation of the temporal stability of depth predictions.

\subsection{Zero-shot depth estimation}
Below we evaluate our method on the mentioned benchmark datasets, comparing it to existing state-of-the-art methods for video depth estimation, including VDA~\cite{chen2025videodepthanything} and FlashDepth~\cite{chou2025flashdepth}.
For all experiments, we evaluate their default models: VDA-L for VDA with window size of 32 frames and the default hybrid model for FlashDepth, which processes frames sequentially.
Additionally, we retrain VDA on our training data and with our hardware constraints for a fair comparison.
Our method uses a window size of 16 frames during inference.
We also include DepthAnythingV2~\cite{yang2024DepthAnythingV2} baseline as our starting point, which processes frames independently.
The core results are summarized in Table~\ref{tab:depth_comparison}, reporting AbsRel, $\delta_1$ as well as the TGM metric.

\begin{table*}[th]
    \centering
    \setlength{\tabcolsep}{1.0pt}
    \scriptsize
    \begin{tabular}{l c ccc|ccc|ccc|ccc| ccc | c}
                                                                            &
                                                                            &
        \multicolumn{3}{c}{\textbf{Infinigen}~\cite{raistrick2023infinite}} &
        \multicolumn{3}{c}{\textbf{Sintel}~\cite{butler2012sintel}}         &
        \multicolumn{3}{c}{\textbf{KITTI}~\cite{geiger2012KITTI}}           &
        \multicolumn{3}{c}{\textbf{TUM RGB-D}~\cite{palazzolo2019bonn}}     &
        \multicolumn{3}{c}{\textbf{Average}}
        \\
        \cmidrule(lr){3-5} \cmidrule(lr){6-8} \cmidrule(lr){9-11} \cmidrule(lr){12-14} \cmidrule(lr){15-17}
        \textbf{Method}                                                     & Data & AbsRel ($\downarrow$) & $\delta_1$ ($\uparrow$) & TGM* ($\downarrow$) & AbsRel ($\downarrow$) & $\delta_1$ ($\uparrow$) & TGM* ($\downarrow$) & AbsRel ($\downarrow$) & $\delta_1$ ($\uparrow$) & TGM* ($\downarrow$) & AbsRel ($\downarrow$) & $\delta_1$ ($\uparrow$) & TGM* ($\downarrow$) & AbsRel ($\downarrow$) & $\delta_1$ ($\uparrow$) & TGM ($\downarrow$) & [ms]
        \\
        \midrule
        DAv2                                                                & 60M  & \underline{0.29}      & \underline{69.32}       & 0.28                & 0.38                  & 57.93                   & 0.89                & \underline{0.13}      & 82.32                   & 0.16                & 0.17                  & 72.81                   & 0.06                & 0.24                  & 70.59                   & 0.35               & 99.2             \\
        FlashDepth                                                          & 500K & 0.32                  & 60.85                   & 0.37                & 0.36                  & 54.84                   & 1.07                & 0.16                  & 74.92                   & 0.21                & \textbf{0.08}         & \textbf{95.00}          & 0.05                & 0.23                  & 71.40                   & 0.43               & 56.3             \\
        VDA                                                                 & 1.3M & \textbf{0.22}         & \textbf{77.71}          & \underline{0.20}    & \textbf{0.27}         & \textbf{67.26}          & \textbf{0.67}       & \textbf{0.08}         & \textbf{94.18}          & \textbf{0.11}       & \textbf{0.08}         & \underline{94.10}       & \textbf{0.02}       & \textbf{0.16}         & \textbf{83.31}          & \textbf{0.25}      & \underline{43.0} \\
        \midrule
        VDA (retrained)                                                     & 360K & \underline{0.29}      & 66.75                   & \textbf{0.19}       & 0.43                  & 57.32                   & 0.77                & \underline{0.13}      & \underline{84.88}       & \underline{0.12}    & \underline{0.11}      & 87.24                   & \textbf{0.02}       & 0.24                  & 74.05                   & 0.28               & \underline{43.0} \\
        StableDPT                                                           & 360K & \underline{0.29}      & 67.22                   & 0.21                & \underline{0.35}      & \underline{60.08}       & \underline{0.73}    & \underline{0.13}      & 83.27                   & 0.13                & 0.12                  & 86.70                   & \underline{0.03}    & \underline{0.22}      & \underline{74.32}       & \underline{0.27}   & \textbf{22.8}    \\
        Relative Improvement                                                &      & 1.0\%                 & 0.7\%                   & -9.4\%              & 17.6\%                & 4.8\%                   & 5.6\%               & -3.9 \%               & -1.9\%                  & -6.5\%              & -2.6\%                & -0.6\%                  & -27.3\%             & 7.5\%                 & 0.4\%                   & 1.1\%              & 47.0\%           \\
        \bottomrule
    \end{tabular}

    \caption{A comparison of depth estimation methods. Best results are in \textbf{bold}, second best are
        \underline{underlined}. *VDA and ours were trained on TGM loss, while FlashDepth and DAv2 were not.
        Since not all data for VDA is public, we report numbers for a retrained same architecture model as our baseline.
        StableDPT achieves competitive performance across all datasets and metrics, while being 2x more efficient during inference.}
    \label{tab:depth_comparison}
\end{table*}

The results across all datasets indicate that our method achieves competitive performance in terms of accuracy and temporal consistency.
Just falling shy of first place in most metrics, our method consistently ranks second best, demonstrating its effectiveness in leveraging temporal information for video depth estimation.

In a more in-depth temporal evaluation of the methods, we also compute additional temporal consistency metrics to highlight the
temporal stability of related work and our proposed method: OPW, TC, TCC, and TMC.
The results are summarized in Table~\ref{tab:temporal_metrics}.
Again, despite fewer data, our method achieves competitive performance across all temporal consistency metrics as well as faster processing.

\begin{table}[ht]
    \centering
    \setlength{\tabcolsep}{1.5pt}
    \scriptsize
    \begin{tabular}{l  ccccccccc}
        \multicolumn{1}{c}{} & \multicolumn{8}{c}{\centering\textbf{Infinigen Dataset}}
        \\
        \cmidrule(lr){2-9}
        \textbf{Metric}      & AbsRel ($\downarrow$)                                    & $\delta_1$ ($\uparrow$) & TC ($\uparrow$)   & OPW ($\downarrow$) & TCC ($\uparrow$)  & TMC ($\uparrow$)  & TGM ($\downarrow$) & R ($\downarrow$)
        \\
        DAv2                 & \underline{0.287}                                        & \underline{69.32}       & \underline{0.820} & 0.039              & 0.724             & \textbf{1.119}    & 0.283              & 2.4              \\
        FlashDepth           & 0.318                                                    & 60.85                   & 0.816             & 0.054              & 0.676             & 1.106             & 0.373              & 3.8              \\
        VDA                  & \textbf{0.218}                                           & \textbf{77.71}          & \textbf{0.827}    & \underline{0.026}  & \textbf{0.826}    & \underline{1.118} & \textbf{0.200}     & \textbf{1.4}     \\
        StableDPT            & 0.292                                                    & 67.22                   & \textbf{0.827}    & \textbf{0.023}     & \underline{0.789} & 1.098             & \underline{0.210}  & \underline{2.2}  \\
        \bottomrule
    \end{tabular}
    \caption{In-depth temporal evaluation of methods on Infinigen dataset on all temporal metrics. StableDPT achieves a competitive performance rank ($\downarrow$) R=2.2 across all metrics.}
    \label{tab:temporal_metrics}
\end{table}

\subsection{Training compute}

\begin{figure}[h]
    \centering
    \includegraphics[width=0.9\linewidth]{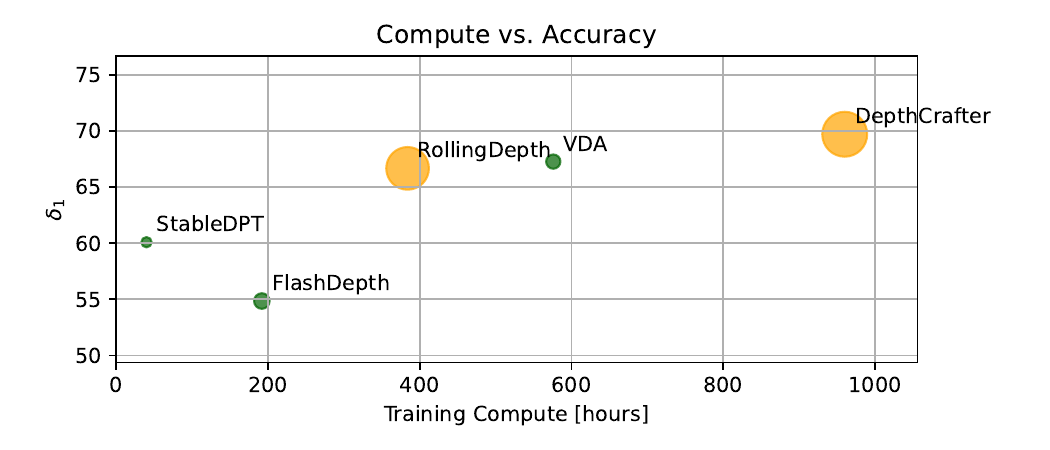}
    \caption{Training compute comparison in GPU hours. Our method requires significantly less compute compared to other methods. Green/orange indicates transformer/diffusion based methods.
        Size indicates inference time.}
    \label{fig:training_compute}
\end{figure}

The training compute of our method is much lower compared to related work in terms of GPU hours.
Our efficient training, leveraging a frozen pre-trained encoder and a lightweight temporal module with temporal-only cross-attention,
enables training on a single RTX 5090 GPU within three days.
This is opposed to multiple high-end GPUs over several weeks required by other methods, see Figure~\ref{fig:training_compute} for a comparison of training compute.

\subsection{Inference time}

We measure the inference time of our method and related works on RTX 5090 GPU and on a sequence of 192 frames, with 518$\times$924 resolution.
It is measured from the moment the video frames are loaded into GPU until all depth maps are computed and processed, this means including all preprocessing and postprocessing steps.
Reported timings exclude warm-up time for the GPU.
The results are summarized in Table~\ref{tab:inference_time}.

\begin{table}[ht]
    \centering
    \setlength{\tabcolsep}{1.5pt}
    \scriptsize
    \begin{tabular}{l | c c c}
        \textbf{Method}  & \textbf{Precision} & \textbf{Per sequence (s)} & \textbf{Per frame (ms)}
        \\
        \midrule
        FlashDepth       & FP16               & 10.8                      & 56.3                    \\
        VDA-L            & FP16               & 8.3                       & 43.0                    \\
        StableDPT (Ours) & FP16               & \textbf{4.4}              & \textbf{22.8}           \\
        \midrule
        FlashDepth*      & FP32               & -                         & -                       \\
        VDA-L            & FP32               & 23.9                      & 124.3                   \\
        StableDPT (Ours) & FP32               & \textbf{12.2}             & \textbf{63.4}           \\
        \bottomrule
    \end{tabular}
    \caption{Inference time comparison on RTX 5090 GPU for a sequence of 192 frames with full and half precision.
        *FlashDepth cannot be run in FP32 precision.}
    \label{tab:inference_time}
\end{table}

\subsection{Qualitative results}

\begin{figure}[h]
    \centering
    \includegraphics[width=1\linewidth]{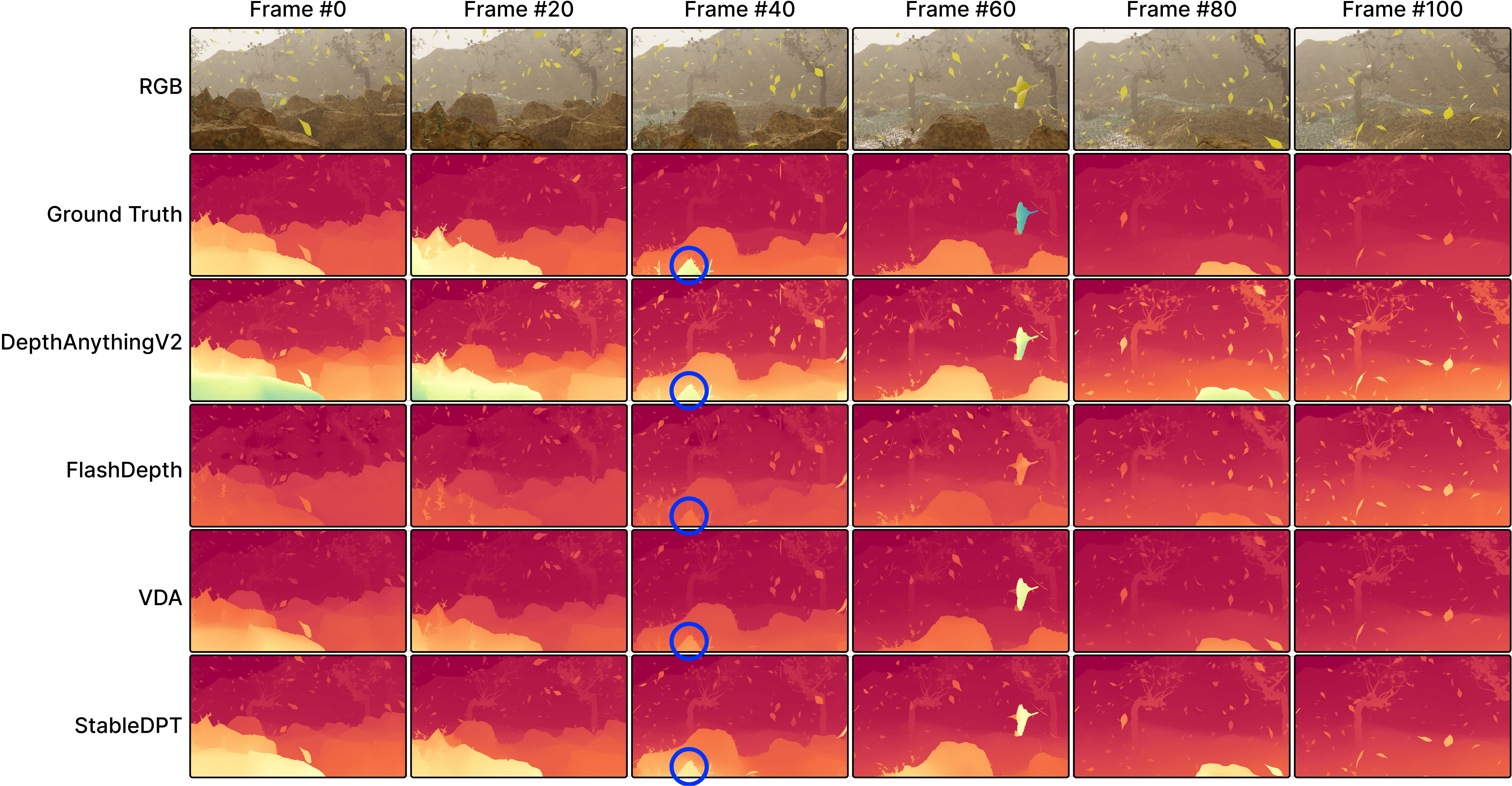}
    \caption{Qualitative comparison of our method with related work on a sequence from Infinigen dataset.}
    \label{fig:qualitative_leaves}
\end{figure}

\begin{figure}[h]
    \centering
    \includegraphics[width=1\linewidth]{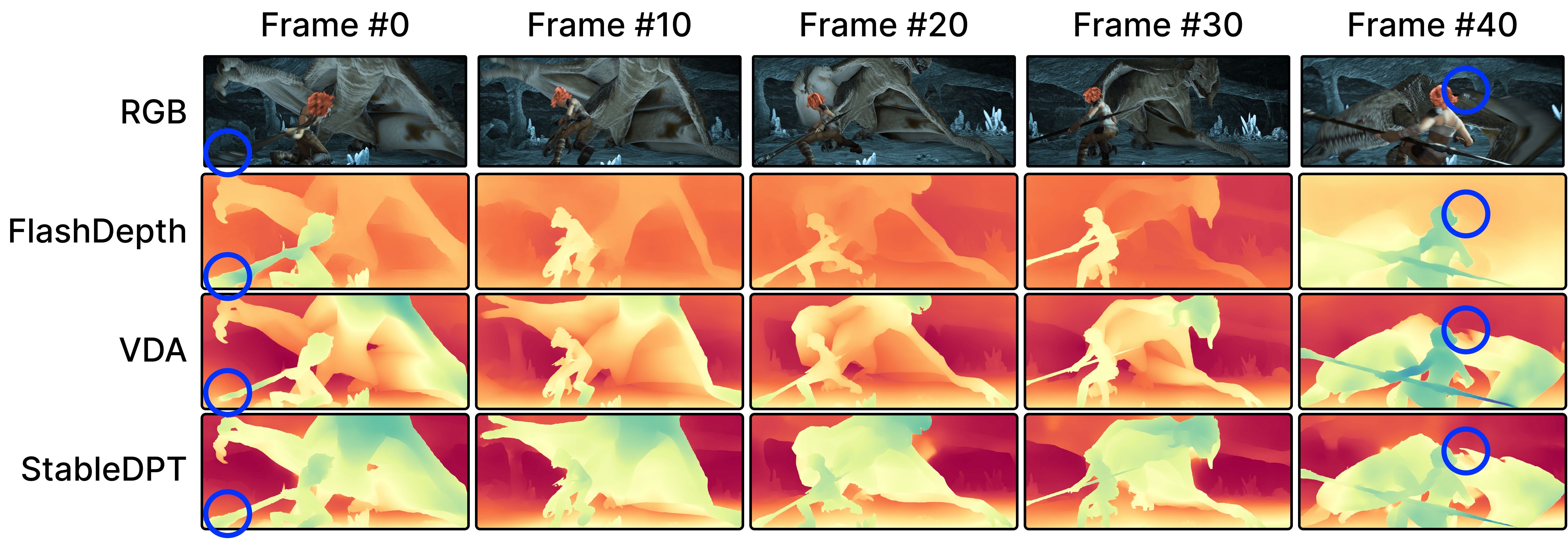}
    \caption{Qualitative comparison of our method with related work on a sequence from Sintel dataset. Blue circle highlights regions with noticeable temporal inconsistencies.}
    \label{fig:qualitative_sintel}
\end{figure}

Here we provide qualitative comparisons of our methods with base models on two sequences, taken from Infinigen and Sintel datasets,
shown in Figures~\ref{fig:qualitative_leaves} and~\ref{fig:qualitative_sintel}.

The qualitative results demonstrate that our method produces temporally consistent depth maps compared to the base model, and highly competitive with VDA.
Our temporal module effectively smooths out these inconsistencies, resulting in more coherent and sharper depth predictions over time,
especially in challenging regions such as object boundaries and fast-moving areas.

\subsection{Ablation: Inference strategies}
We compare different inference strategies for processing videos of arbitrary length, mentioned in Section~\ref{sec:inference_strategy}.
We define and evaluate on four different strategies.
\textbf{Batched~(B)} inference splits the video into non-overlapping batches of fixed size, and each batch is processed independently.
\textbf{VDA} inference processes the video in overlapping windows. Overlapping frames are averaged to produce the final depth map.
\textbf{Strided~(S)} inference processes the video using a strided approach, but without the global context.
\textbf{Strided~+~Keyframe~(S+KF) (proposed)}, same as \textit{Strided}, but with cross-attending to global keyframes.
All methods use the same trained model for evaluation and were run with window size of eight frames.
Our method is far more efficient than other methods (2x), since it avoids redundant wasteful computations, see Table~\ref{tab:inference_time}.
Finally, as shown in Table~\ref{tab:inference_strategies}, our proposed inference strategy outperforms all other strategies in terms of temporal consistency.

\begin{table}[ht]
    \centering
    \setlength{\tabcolsep}{1.0pt}
    \scriptsize
    \begin{tabular}{l | cccccccc}
        \toprule
        \textbf{Strategy} & AbsRel ($\downarrow$) & $\delta_1$ ($\uparrow$) & TC ($\uparrow$)   & OPW ($\downarrow$) & TCC ($\uparrow$)  & TMC ($\uparrow$)  & TGM ($\downarrow$) & R ($\downarrow$)
        \\
        \midrule
        B                 & \underline{0.292}     & 66.450                  & 0.824             & \underline{0.025}  & 0.785             & 1.078             & 0.228              & 3.1              \\
        VDA               & \textbf{0.290}        & 66.260                  & \underline{0.827} & \textbf{0.022}     & \underline{0.791} & 1.090             & \underline{0.210}  & \underline{2.3}  \\
        S                 & 0.294                 & \textbf{67.243}         & 0.823             & 0.033              & 0.740             & \underline{1.096} & 0.271              & 3.4              \\
        S+KF              & 0.294                 & \underline{66.849}      & \textbf{0.828}    & \textbf{0.022}     & \textbf{0.794}    & \textbf{1.100}    & \textbf{0.208}     & \textbf{1.6}     \\
        \bottomrule
    \end{tabular}
    \caption{A comparison of inference strategies on Infinigen dataset.}
    \label{tab:inference_strategies}
\end{table}

\begin{figure}
    \centering
    \includegraphics[width=1\linewidth]{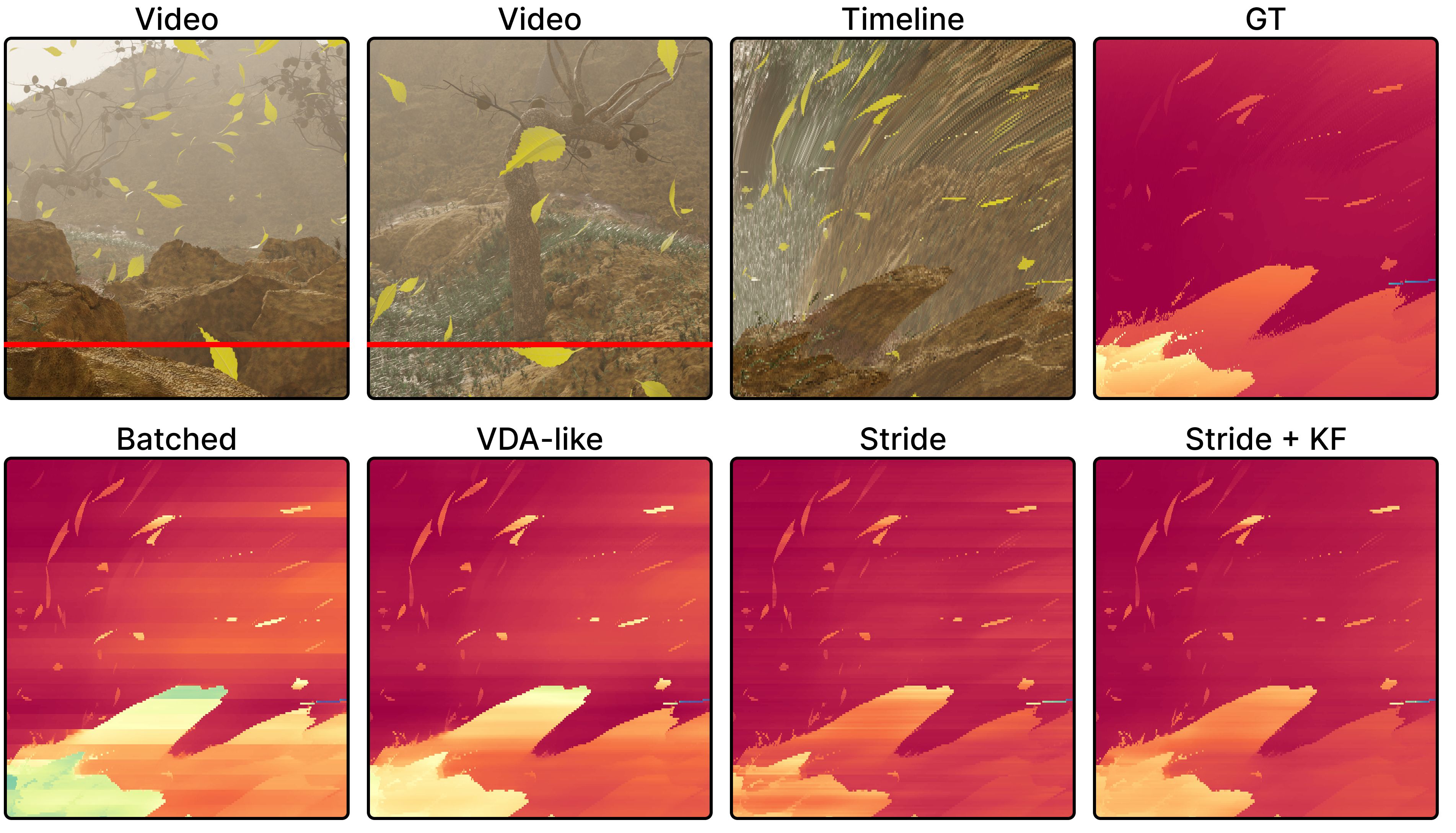}
    \caption{X-t slice visualization of different inference strategies.
        StableDPT with keyframes (S+KF) produces the most consistent results over time and does not have blurry or blocky regions, characteristic of VDA-like strategy.}
    \label{fig:xt_slice}
\end{figure}

We also propose an insightful visual comparison method (see Figure~\ref{fig:xt_slice}) for inference strategies by slicing the video over time.
In this visualization, one pixel row is taken from each consecutive frame in the video,
creating an image that represents how depth values change over time for a specific horizontal line in the video.
This allows us to visually assess the temporal consistency of depth predictions across different inference strategies.

\subsection{Ablation: Temporal module design}

In order to fit the training onto a consumer-grade GPU, we ablated the head architecture and reduced the number of temporal modules and attention layers to two each.
This allows training at reduced memory, compute and improves accuracy, see supplemental for details.

\subsection{Ablation: Long video depth drift}

Due to the global context provided by keyframes during inference, our method is less prone to depth drift over long sequences.
In the supplemental material, we provide an in-depth evaluation of depth drift, showing plots of depth accuracy over time.

%% file: sec/6_conclusion.tex
\section{Conclusion and outlook}

In this work, we presented a novel approach to adapt single image models for video estimation by adding temporal transformer modules with cross-attention mechanisms.
Our method successfully captures global video context and inter-frame relationships leading to more stable and coherent predictions.
While the main application is depth prediction, please note that the proposed temporal modules are general and could be applied to other dense prediction tasks, \eg~optical flow or semantic segmentation.

Three key benefits of our approach are its flexibility, efficiency and end-to-end training capability.
Our temporal-only modules can be easily integrated into existing single image models without requiring extensive modifications.
The strided inference strategy we proposed allows for processing videos of arbitrary length while maintaining manageable computational and memory requirements, \eg~2x more efficiency than VDA.

Our evaluations on multiple benchmark datasets demonstrate the improved temporal consistency and competitive state-of-the-art performance.
This performance is reached despite limited video training data and computing resources.
Future work could explore the real-time application of our method, as well as its extension to other dense prediction tasks beyond depth estimation.
Overall, our approach represents a significant step towards more robust, efficient and general transformer-based video analysis and understanding.